\pdfoutput=1

\documentclass[11pt]{article}

\PassOptionsToPackage{table}{xcolor}
\usepackage{ACL2023}

\usepackage{times}
\usepackage{latexsym}
\usepackage{booktabs}
\usepackage[table]{xcolor}
\usepackage{hyperref}
\usepackage{svg}
\usepackage{amsmath}

\usepackage[T1]{fontenc}

\usepackage[utf8]{inputenc}

\usepackage{microtype}

\usepackage{inconsolata}

\definecolor{light}{RGB}{255,127,0}
\definecolor{lightgreen}{RGB}{255,127,0}
\definecolor{green}{RGB}{83,199,74}
\definecolor{darkgreen}{RGB}{255,127,0}
\definecolor{dark}{RGB}{255,127,0}

\title{SLABERT Talk Pretty One Day: Modeling Second Language Acquisition with BERT}

\author{\bf $^\star$Aditya Yadavalli$^1$ \quad
\bf $^\star$Alekhya Yadavalli$^2$ \quad
\bf Vera Tobin$^2$ \\
$^1$ Karya Inc.\thanks{\ \ Work done as a visiting researcher at Case Western Reserve University} \quad \\
$^2$ Language and Cognition Lab, Case Western Reserve University \\
\tt aditya@karya.in \quad
\tt \{alekhya.yadavalli, vera.tobin\}@case.edu \\
}

\usepackage{lipsum}

\newcommand\blfootnote[1]{%
  \begingroup
  \renewcommand\thefootnote{}\footnote{#1}%
  \addtocounter{footnote}{-1}%
  \endgroup
}

\begin{document}

\maketitle
\begin{abstract}
Second language acquisition (SLA) research has extensively studied cross-linguistic transfer, the influence of linguistic structure of a speaker's native language [L1] on the successful acquisition of a foreign language [L2]. Effects of such transfer can be positive (facilitating acquisition) or negative (impeding acquisition). We find that NLP literature has not given enough attention to the phenomenon of \textit{negative transfer}. To understand patterns of both positive and negative transfer between L1 and L2, we model sequential second language acquisition in LMs. Further, we build a Mutlilingual Age Ordered CHILDES (MAO-CHILDES)---a dataset consisting of 5 typologically diverse languages, i.e., German, French, Polish, Indonesian, and Japanese---to understand the degree to which native Child-Directed Speech (CDS) [L1] can help or conflict with English language acquisition [L2]. To examine the impact of native CDS, we use the TILT-based cross lingual transfer learning approach established by \citet{music} and find that, as in human SLA, language family distance predicts more negative transfer. Additionally, we find that conversational speech data shows greater facilitation for language acquisition than scripted speech data. Our findings call for further research using our novel Transformer-based SLA models and we would like to encourage it by releasing our code, data, and models.\blfootnote{Code, data and models are here: \url{https://github.com/AdityaYadavalli1/SLABERT}}
\blfootnote{$\star$ Authors contributed equally}
\end{abstract}

\section{Introduction}
Cross-linguistic transfer can be described as the influence of native language [L1] properties on a speaker’s linguistic performance in a new, foreign language [L2]. The interaction of the linguistic structure of a speaker’s L1 with the successful acquisition of L2 results in what are termed as \textit{transfer effects}. Transfer effects appear in various aspects of linguistic performance, including vocabulary, pronunciation, and grammar \cite{Jarvis2007CrosslinguisticII}. Cross-linguistic transfer can be positive or negative in nature: positive transfer refers to the facilitating effects of one language in acquiring another (e.g., of Spanish vocabulary in acquiring French) and \textit{negative transfer} between the learner's native [L1] and target [L2] languages, producing errors. The greater the differences between two languages, the greater the negative effects.

While cross-lingual transfer has received considerable attention in NLP research \cite{wu-dredze-2019-beto, Wu2019EmergingCS, Conneau2017WordTW, conneau-etal-2018-xnli, artetxe-etal-2018-robust, Ruder2017ASO}, most of this research has concentrated on practical implications such as the degree to which the right tokenizer can optimize cross-lingual transfer, and has not looked at the kind of sequential transfer relationships that arise in human second language acquisition. Meanwhile, approaches like the Test for Inductive Bias via Language Model Transfer (TILT) \cite {music} focus on positive transfer with divergent pairs of training sets, such as MIDI music and Spanish, to shed light on which kinds of data induce generalizable structural features that linguistic and non-linguistic data share. Patterns of both positive and negative transfer between a given L1 and L2, however, can be a valuable source of information about general processes of second language acquisition and typological relationships between the languages in question \cite{Berzak2014ReconstructingNL}. 

Most cross-lingual models do not mimic how humans acquire language, and modeling the differences between first and second language acquisition is a particularly under-explored area. To engage with questions about second language acquisition using LMs, we model sequential second language acquisition in order to look more closely at both positive and negative transfer effects that may occur during the acquisition of L2. 

Using Child-Directed Speech (CDS) to create L1 training sets that are naturalistic, ecologically valid, and fine-tuned for language acquisition, we model the kind of cross-linguistic transfer effects that cause linguistic structure of the native L1 to influence L2 language acquisition in our novel Second Language Acquisition BERT (SLABERT) framework. The resulting models, when tested on the BLiMP (Benchmark of Linguistic Minimal Pairs for English) grammar test suite \cite{warstadt-etal-2020-blimp-benchmark}, show that L1 may not only facilitate L2 learning, but can also interfere. To the extent that interference is considered in NLP research, it is often understood simply as a failure of positive transfer in model training. We suggest, instead, that these results should be analyzed in terms of distinctive patterns of both negative and positive transfer, which can reveal not just the existence of generalizable features across datasets, but also finer-grained information about structural features of these languages and their accessibility to second language learners.

\section{Related Work}

\label{sec:rel_work}

Our work is closely related to and in many ways builds on the work done by \citet{huebner-etal-2021-babyberta}. They proposed that Child-Directed Speech has greater potential than other kinds of linguistic data to provide the structure necessary for language acquisition, and released BabyBERTa, a smaller sized RoBERTa \cite{Liu2019RoBERTaAR} model designed to investigate the language acquisition ability of Transformer-based Language Models (TLM) when given the same amount of data as children aged 1-6 get from their surroundings. They also released Zorro, a grammar test suite, that is compatible with the small vocabulary of child-directed input. 

Child-directed speech (CDS) refers to the special register adopted by some adults, especially parents, when talking to young children \cite{Saxton2009TheIO}. CDS typically features higher fundamental pitch, exaggerated intonation, slower speech, and longer pauses than Adult-Directed Speech (ADS) \cite{clark_2016}. Utterances in CDS are usually well-formed grammatically, but are syntactically simpler than ADS, often comprising single word utterances or short declaratives. Adults often repeat words, phrases, and whole utterances in CDS \cite{turkishlangacq, 10.2307/1127555} and make fewer errors \cite{verbalenv} than they do in ADS. CDS also tends to use a smaller and simplified vocabulary, especially with very young children \cite{hayes_ahrens_1988}. While the universality and necessity of CDS for language acquisition is a matter of debate \cite{pinker_1995, hornstein_nunes_grohmann_2005, HAGGAN200217}, it is likely that the features of CDS are universally beneficial in language acquisition \cite{Saxton2009TheIO}. NLP literature suggests that are certain benefits when models are trained on CDS \cite{Gelderloos2020LearningTU}. Studies from other fields suggest that the pitch contours, repetitiveness, fluency, and rhythms of CDS make it easier for children to segment speech, acquire constructions, and understand language \cite{Cristia2011FinegrainedVI, thiessen2005, KemlerNelson1986HowTP, Ma2011WordLI, Soderstrom2008AcousticalCA, Kirchhoff2003StatisticalPO}. Many of these distinctive qualities of CDS seem tailor-made for human language acquisition, which is why we use CDS data as L1 in our SLABERT models.

Several recent studies confirm that the distinctive distributional features of CDS influence the grammatical and lexical categories that children acquire. For instance, \citet{Mintz2003FrequentFA} found that "frequent frames" in CDS--commonly recurring co-occurance patterns of words in sentences--yield very accurate grammatical category information for both adults and children. Similarly, \citet{verb} found that patterns of frequent use and, importantly, reinforcement in CDS-specific conversational exchanges were most predictive of the constructions children learn. Together, these findings suggest that both token distribution and the distinctive conversational structure of CDS provide useful reinforcement for acquisition. Therefore, when training our L1 model, we pay attention to qualities of the training input such as the conversational structure. 

In second language acquisition (SLA) research, patterns of negative transfer are a topic of much interest and have been considered a source of information both about what happens in second language learning and what it can reveal about the typological relationships between L1 and L2. For instance, \citet{Dulay1974ErrorsAS} show that closely analyzing data from children learning a second language reveals that some errors are due to L1 interference (\textit{negative transfer}), while others arise from developmental cognitive strategies similar to those made during L1 acquisition (\textit{developmental errors}). \citet{Berzak2014ReconstructingNL} show a strong correlation between language similarities derived from the structure of English as Second Language (ESL) texts and equivalent similarities obtained directly from the typological features of the native languages. This finding was then leveraged to recover native language typological similarity from ESL texts and perform prediction of typological features in an unsupervised fashion with respect to the target languages, showing that structural transfer in ESL texts can serve as valuable data about typological facts.

The phenomenon of cross-linguistic transfer has received considerable attention in NLP research in the context of multilingual Language Models \cite{wu-dredze-2019-beto, Wu2019EmergingCS, Conneau2017WordTW, conneau-etal-2018-xnli, artetxe-etal-2018-robust, Ruder2017ASO}. Our investigation is particularly inspired by \citet{music}'s Test for Inductive Bias via Language Model Transfer (TILT). This is a novel transfer mechanism where the model is initially pre-trained on training data [L1]. Next, they freeze a part of the model and fine-tune the model on L2. Finally, they test the resulting model on a test set of L2. We follow a similar approach to our model's second language acquisition. 

\begin{table*}[t]
\small
\centering
\begin{tabular}{lccccc}
\toprule
\textbf{Language} &\textbf{Vocabulary} & \textbf{Total tokens} & \textbf{Avg. Sentence Length} & \textbf{No. of Children} & \textbf{Utterances}\\
\midrule
{American English} & {27,723} & {4,960,141} & {5.54832} & {1117} & {893,989}\\ 
{French} & {22,809}  & {2,473,989} & {5.74531} & {535} & {487,156}\\
{German} & {59,048} & {4,795,075} & {5.65909} & {134} & {951,559}\\
{Indonesian} & {21,478} & {2,122,374} & {3.97058} & {9} & {572,581}\\ 
{Polish} & {31,462} & {493,298} & {5.84276} & {128} & {84,578}\\ 
{Japanese} & {44,789} & {2,397,386} & {4.17552} & {136} & {588,456}\\
{Wikipedia-4} & {84,231} & {1,907,706} & {23.8456} & {-} & {80,000} \\
{English ADS} & {55,673} & {905,378} & {13.1901} & {-} & {74,252}\\
\bottomrule
\end{tabular}
\caption{MAO-CHILDES corpus statistics: the number of unique tokens, total tokens, the average sentence length, the total number of children, and the mean age of child for each language dataset is presented}
\label{tab:lang_stats}
\end{table*}

\section{Data}

\subsection{Why Child-Directed Speech}
We wanted L1 training sets that are both realistic and fine-tuned to teach language to developmental (first language) learners. We also wanted to reproduce the findings of \citet{huebner-etal-2021-babyberta} which suggest that Child-Directed Speech as training data has superior structure-teaching abilities for models compared to scripted adult-directed language.

The BabyBERTa studies \cite{huebner-etal-2021-babyberta} found that their LM required less data than RoBERTa to achieve similar (or greater) linguistic/syntactic expertise (as tested by Zorro), and suggested that CDS is better than Wikipedia text for teaching linguistic structure to models. Given these findings and widespread support in cognitive science and linguistics for the facilitative nature of CDS in child language learning, we choose to use CDS data from five different languages as our L1s to examine our hypothesis that preexisting linguistic structure of L1 interacts differentially with the acquisition of L2 (English). 

Additionally, building on the \citet{huebner-etal-2021-babyberta} efforts to find superior training data for LMs in general, we explore the possibility that comparing conversational CDS with scripted ADS is a less fair comparison than comparing the quality of conversational CDS with that of conversational ADS as training input for LMs. 

\subsubsection{Why CHILDES}

Our focus in training the Child-Directed Speech model is on replicating for the LM, as closely as possible, the primary linguistic input of young children. While young children are exposed to passive Adult-Directed Speech, speech that is directed at them and intended to communicate with them plays a more central role in the child’s linguistic experience \cite{Soderstrom2007BeyondBR}. For this reason, we use a language database of naturalistic speech directed at children. The CHILDES \cite{CHILDES} database, a component of the larger TalkBank corpus, is a vast repository of transcriptions of spontaneous interactions and conversations between children of varying ages and adults.\footnote{\url{https://talkbank.org}} The database comprises more than 130 corpora from over 40 different languages and includes speech directed at children from ages of 6 months to 7 years. The large selection of languages permits us the necessary flexibility in choosing different languages for our L1 data (see Section \ref{sec:lang_sel} for more on Language Selection). The range of child ages allows us to train our models with increasingly complex linguistic input, emulating the linguistic experience of a growing child.

\subsubsection{Language Selection}
\label{sec:lang_sel}

Our focus is on cross-linguistic transfer of language structure; therefore, we use a simple selection criterion and choose five languages with varying distance from English according to their language family: German, French, Polish, Indonesian, and Japanese. We hypothesize languages that are structurally similar to English should perform better (show more positive transfer and less negative transfer). German, French, and Polish, like English, are all Indo-European languages. However, each of these languages belongs to a unique genus: German and English are Germanic languages, French is a Romance language, and Polish is a Slavic language. While English and French do not share the same genus, there is much overlap between the two languages due to the substantial influence of French on English stretching back to the time of Norman Conquest. Japanese belongs to the Japanese language family and Indonesian to the Austronesian language family.

\subsubsection{Using the AO-CHILDES corpus}

The AO-CHILDES (AO: age-ordered) corpus was created from \citet{inbook} American English transcripts from the CHILDES database. To curate the American English collection, we followed the same cleaning criteria as \citet{inbook}: only transcripts involving children 0 to 6 years of age were procured, from which child (non-adult) utterances and empty utterances were omitted. The initial CHILDES transcriptions were converted from CHAT transcription format to csv format files using \texttt{childes-db} \cite{childes-db} to conduct the data cleaning processes. The resulting dataset, which now contains 2,000,352 sentences, 27723 unique words, and 4,960,141 total word tokens, forms the American English input. This cleaning process was repeated for the corpora of German, French, Polish, Japanese, and Indonesian to create the dataset for each language (see Table \ref{tab:lang_stats} for the language statistics).

\subsubsection{MAO-CHILDES}

For the sake of simplicity we refer to the corpus resulting from the collective datasets of the six languages as MAO-CHILDES (MAO is short for Multilingual Age-Ordered) to show that the transcripts it contains include a selection of different languages and also are ordered by age of child (see Table \ref{tab:lang_stats}). 

Data in MAO-CHILDES is not uniformly distributed across languages, as seen in Table \ref{tab:lang_stats}. First, Polish is represented by significantly less data than every other language. Second, Indonesian has a lower number of unique tokens compared to other languages. The Indonesian data is also only collected from conversations with 9 children, a much smaller sample size compared to the other languages, which have sample sizes in the hundreds if not thousands. Third, the average sentence length of the Asian languages---Indonesian and Japanese---is smaller than any of the other languages. The effect of these variations in data, caused by both available resources and natural linguistic characteristics of the languages, on the performance of the cross-lingual model is anticipated.

\begin{figure*}[t]
\centering
\includegraphics[width=\textwidth]{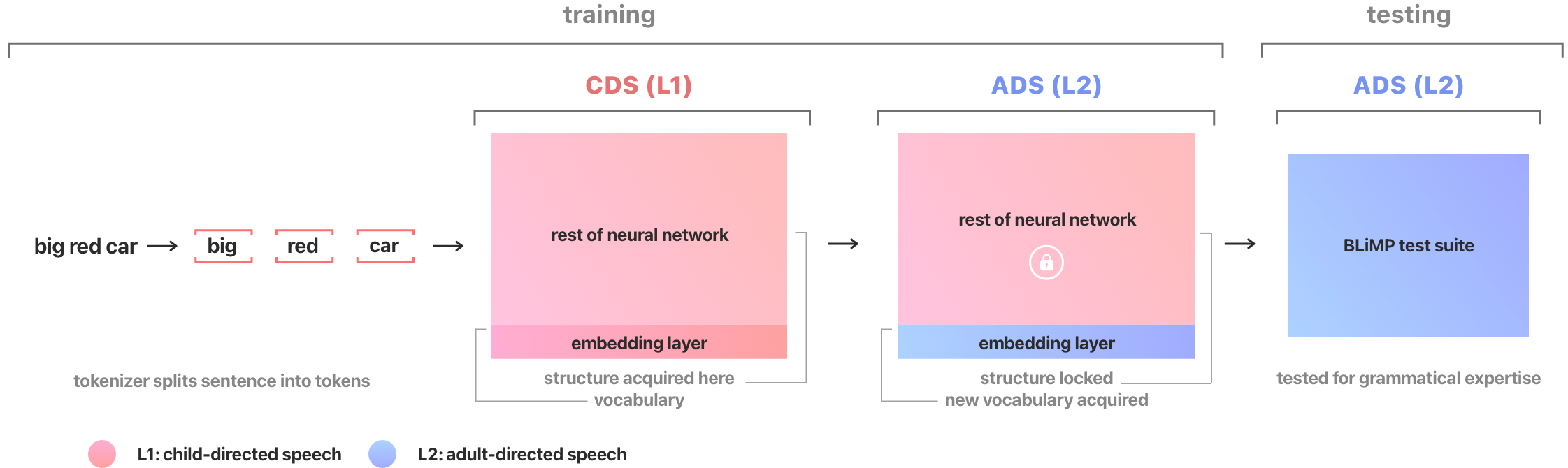}
\caption{Diagram illustrating our experimental process for each L1, as listed in Table \ref{tab:lang_stats}. Training occurs in two stages and each model is finally tested on the BLiMP test suite.}
\label{fig:neural_net}
\end{figure*}

\subsection{Adult-Directed Speech corpus}

The Adult-Directed Speech (ADS) corpus comprises conversational speech data and scripted speech data. We build on the BabyBERTa efforts to find superior training data for LMs (in general) by experimenting with conversational ADS and comparing its training utility with that of conversational CDS. This investigation is aimed at narrowing down the true source, child-directed language or conversational language, of the reduced data size requirements of BabyBERTa.

To create our conversational ADS corpus, we use the sample COCA SPOKEN corpus.\footnote{\url{https://www.corpusdata.org}} COCA (Corpus of Contemporary American English) is one of the most widely used corpora of English for its rich representation of texts from a wide range of genres, dialects, and time periods. The SPOKEN genre comprises transcriptions of spontaneous conversations between adults. To clean this sample corpus we followed a three step process: 
\begin{itemize}
    \item All spoken disfluencies such as pauses, laughter, and filler utterances encoded in the spoken transcripts were cleaned.
    \item All meta tags that mention the names of the speakers were removed. 
    \item Finally, the data was sampled manually to check that the corpus was clean. 
\end{itemize}

After cleaning, we were left with 74,252 utterances. We use this cleaned corpus to train our conversational Adult-Directed Speech (ADS) model. 

To replicate the findings of the BabyBERTa study, we also train a model on scripted ADS. To create our scripted ADS corpus, we randomly sample 80,000 sentences from Wikipedia-3 \cite{huebner-etal-2021-babyberta}, which we term Wikipedia-4, so that the data size of conversational ADS and scripted ADS is approximately equal, to allow fair comparison. All the information about the data we used is in Table \ref{tab:lang_stats}. 

\section{Experimental Setup}
\label{sec:exp_set}

We use BabyBERTa \cite{huebner-etal-2021-babyberta} to run all our experiments. BabyBERTa is a smaller-sized RoBERTa \cite{Liu2019RoBERTaAR} tuned to perform well on data of the size of AO-CHILDES. However, we make additional changes to the vocabulary size of the model as we found that to improve the results of the model. The implementation details of the model can be found in Appendix \ref{sec:imp_details}. 

We follow the TILT approach introduced by \citet{music} to originally test the LSTM-based \cite{lstm} LM’s structure acquisition. Their general approach is followed in the current study with a few notable changes (See Figure \ref{fig:neural_net}). Our approach comprises two stages: (1) train the model on L1 (CDS language) (2) freeze all parameters except the word embeddings at the transfer stage of the experiment, and fine-tune the model on L2 (English ADS). Finally, the resulting model is tested on a test set of L2 for which we use the Benchmark of Linguistic Minimal Pairs (BLiMP) \cite{warstadt-etal-2020-blimp-benchmark}, a challenge set for evaluating the linguistic knowledge of the model on major grammatical phenomena in English. Our study deviates from \citet{music} approach in three ways: (1) instead of using LSTM-based LMs we use Transformer-based LMs \cite{Vaswani2017AttentionIA} (2) they freeze all layers except the word embedding and linear layers between the LSTM layers however, for simplicity we freeze all parameters except the word embeddings (3) while they report their findings based on LM perplexity scores, we use the BLiMP test suite to report how L1 structure (particularly, syntax and semantics) affects L2 acquisition in our Transformer-based LMs.

\begin{figure*}[t]
\centering
\includegraphics[width=\textwidth]{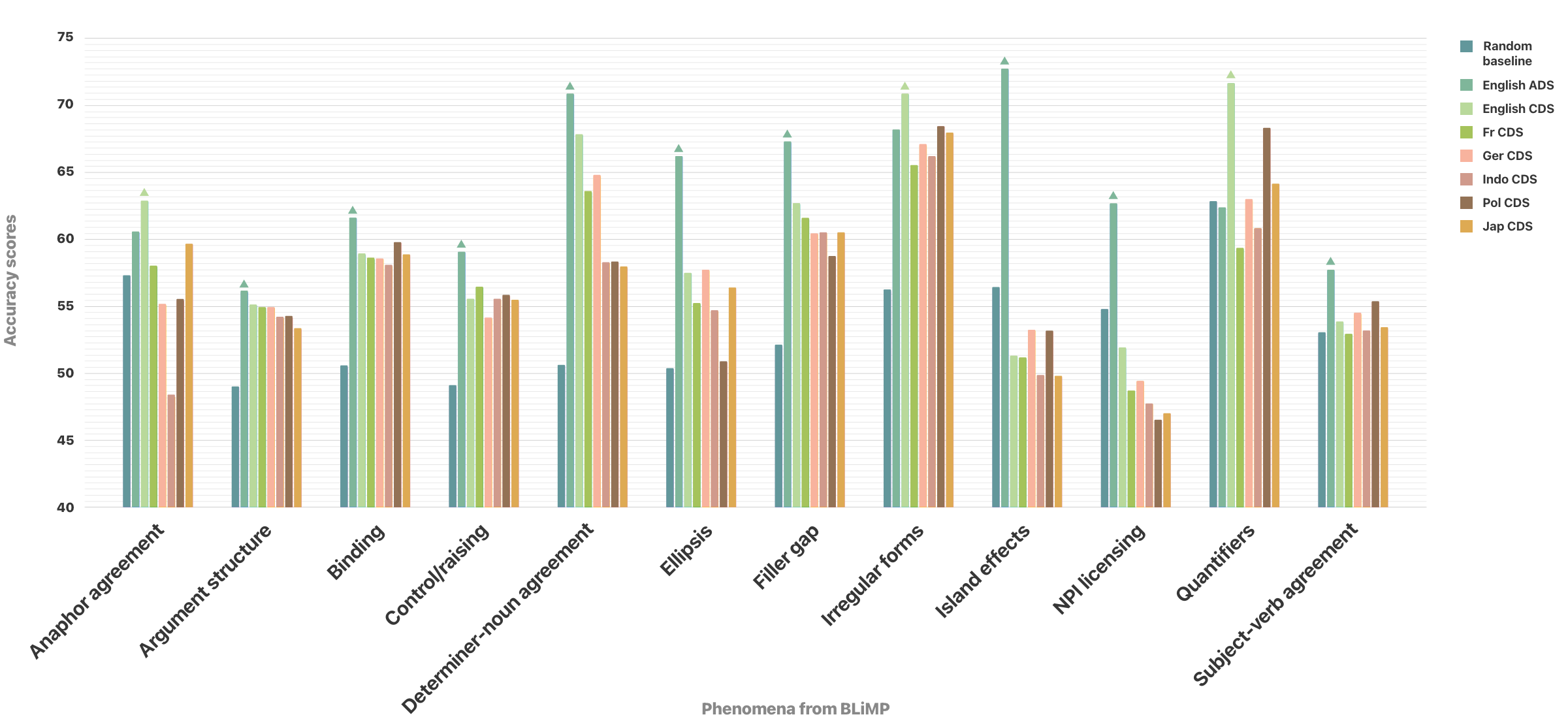}
\caption{Performance of model on various grammatical phenomena from the BLiMP test suite}
\label{fig:results}
\end{figure*}

There are two experiments for which we follow a different procedure than what is explained above:
\begin{itemize}
    \item In the case of \texttt{random-baseline} experiment, we freeze all of the model except the embeddings and let the model train on conversational English ADS. The corresponding tokenizer is also trained on conversational English ADS. This experiment is run in order to have the right benchmark to compare against. This method prevents the model from picking up any grammatical structure from the training data, while allowing it to acquire English vocabulary. 
    \item In the case of the scripted ADS and conversational ADS experiments, we do not employ TILT-based cross lingual transfer. We train the model from scratch on scripted ADS and conversational ADS respectively. 
\end{itemize}

\textbf{Testing:} We use the BLiMP grammar test suite to evaluate the linguistic knowledge of our model. BLiMP consists of 67 paradigms categorized into 12 major grammatical phenomena in English. Each of these 67 datasets comprises 1,000 minimal pairs i.e. pairs of minimally different sentences, one of which is grammatically acceptable and the other not (refer to \citet{warstadt-etal-2020-blimp-benchmark} for a detailed description of the test suite).

\section{Results and Discussion}
\label{sec:res}

\subsection{Results}
\label{results}
The proportion of the BLiMP minimal pairs in which the model assigns a higher probability to the acceptable sentence informs the accuracy of the model. A total of 9 models are compared in their performance using the accuracy scores obtained on 12 different grammatical tests from the BliMP test suite. We report the results for all models in Figure \ref{fig:results} (see Appendix \ref{sec:comp_results} for detailed results). The model trained on conversational English ADS achieves the highest accuracy and the one trained on Indonesian CDS achieves the lowest. Despite the conversational English ADS corpus size being at least 10x smaller than the CDS corpora sizes, it performs the best in 9 out of 12 grammatical phenomena from the BLiMP test suite. CDS demonstrates higher accuracy only in anaphor agreement, irregular forms, and quantifiers. Overall, English CDS performs 5.13 points behind English ADS. These results show that (conversational) Adult-Directed speech makes for superior training data for models as compared to (conversational) Child-Directed Speech. From Figure \ref{fig:results}, we note a few other significant trends:

First, the results indicate that conversational speech data form a superior training data for language models in general as compared to the conventional scripted data. Table \ref{tab:lang_results} compares the performance of models when trained on different types of training inputs of the same language (English): scripted ADS (Wikipedia-4), conversational ADS, and conversational CDS. Among the three, the performance of the model trained on conversational ADS is highest, followed by conversational CDS, and lastly scripted ADS. Important to note here is that, corroborating the findings of the BabyBERTa study, conversational CDS still outperforms scripted ADS (Wikipedia-4) but falls behind compared to conversational ADS. These results suggest that conversational speech data are a more effective training source for models than scripted data (more on this in Section \ref{conversational_better}).

Second, the results show a negative correlation between the distance of the CDS language from English and the performance of the model, i.e., as the typological distance between L1 and L2 increases, the performance of the model decreases. We term this the Language Effect. This finding supports our hypothesis that, given the relation between transfer errors and typological distance between L1 and L2 \cite{ringbom}, the increasing structural dissimilarities between the L1 (CDS language) and the L2 (always English ADS) should adversely impact the performance of the model (more on this in Section \ref{language_effect}).

Third, the results show that CDS performs worse than ADS in several grammatical phenomena (9 out of 12). Considering the simplistic and facilitating structure and, more importantly, the ecologically valid nature of CDS, these results engender some interesting hypotheses which we discuss briefly in Section \ref{CDS_failure}.

Fourth, we see several results in which individual models perform poorly on individual tests in ways that are not cleanly predicted by general trends. We believe these results reflect patterns of negative transfer, in which L1-specific structures actively interfere with the acquisition of structures in L2 (more on this in Section \ref{sec:not_expected}).

\subsection{Conversational vs. Scripted Data}
\label{conversational_better}
The conventional training data for LMs is scripted adult-directed speech, perhaps owing to its easily accessible nature compared to other forms of data, such as conversational ADS or any form of CDS. However, our findings demonstrate that conversational data yields better model performance than scripted data (see Table \ref{tab:lang_results}). The best accuracy scores are produced by conversational ADS on 67\% of the phenomena, by conversational CDS on 25\% of the phenomena, by scripted ADS on 8\% of the phenomena. Conversational data may make for a better training input for language acquisition given a higher level of interactive components in its composition which is an essential feature of language acquisition in children. Much of the previous research has looked at what conversational language does for the people who are directly contributing to the conversation in question. For instance, there is a general tendency for speakers to reproduce grammatical \cite{Bock1986SyntacticPI, Gries2005SyntacticPA} elements of their interloctor's previous utterances. These behaviors both enhance interactive alignment \cite{Bois2014TowardsAD} and ease cognitive load for utterance planning \cite{Bock1986SyntacticPI, Pickering2008StructuralPA}. Studies of children's conversational behavior \cite{verb, Kymen2014DialogicSA} show, similarly, that children use their interlocutors' immediately preceding utterances as resources for producing and reinforcing construction types they are in the process of acquiring. Our findings suggest that the resulting distributional patterns of "dialogic syntax" \cite{Bois2014TowardsAD} in the conversational record leave a trace that can make conversational data especially informative for model training.

\subsection{Language Effect}
\label{language_effect}

\begin{figure}[h]
    \centering
    \includegraphics[width=0.5\textwidth]{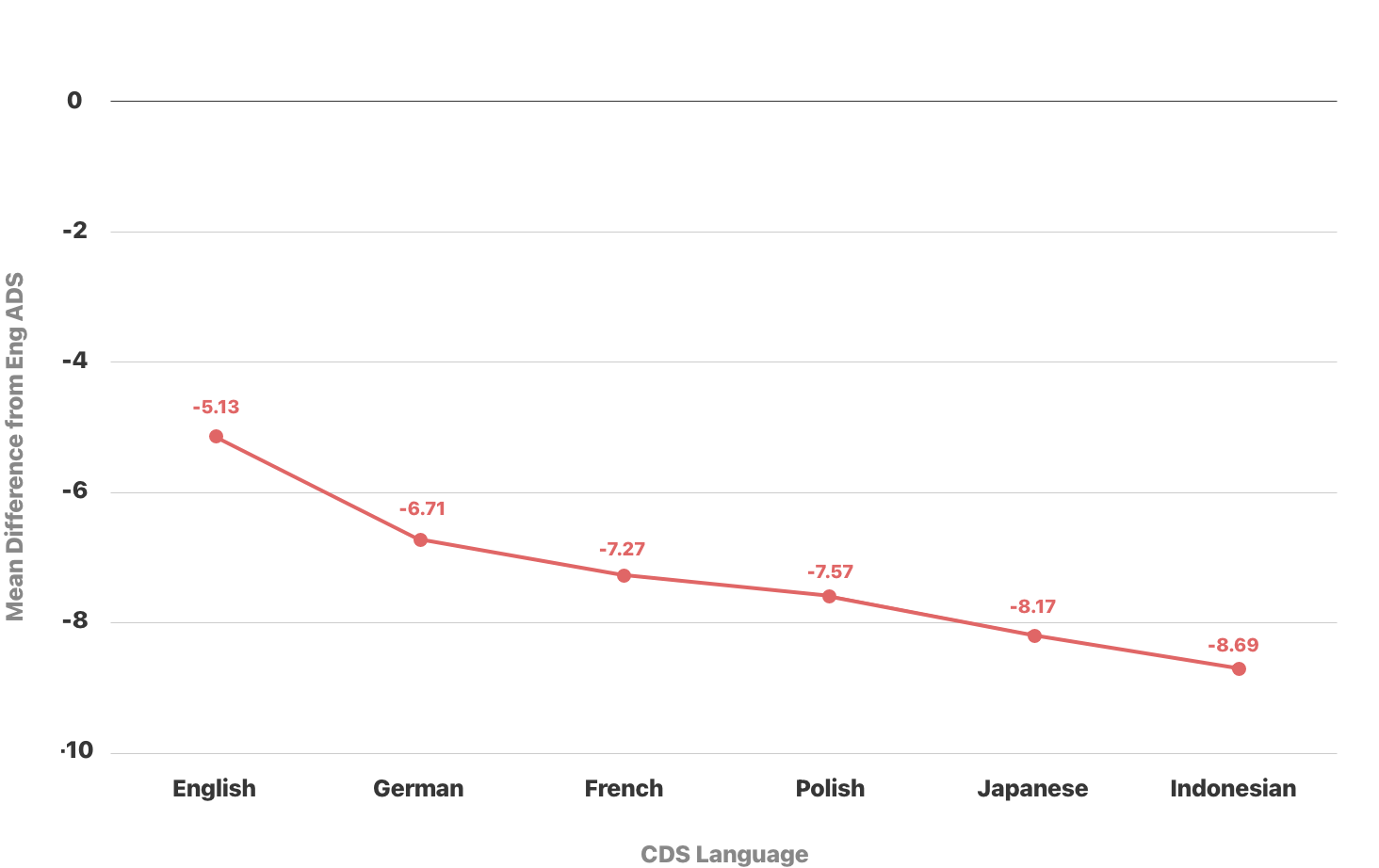}
    \caption{Mean multilingual CDS performance compared to ADS}
    \label{fig:real_language_effect}
\end{figure}

\begin{table*}[h!]
\small
\centering
\begin{tabular}{lccc}
\toprule
\textbf{Phenomenon} &\textbf{Wikipedia-4} & \textbf{Conversational ADS} & \textbf{Conversational CDS} \\
\midrule
{Ananaphor Agreement} & \cellcolor{green!25}51.4 &	\cellcolor{green!50}60.6 & \cellcolor{green!50}62.9 \\
{Argument Structure} & \cellcolor{green!25}54.5	& \cellcolor{green!25}56.1 & \cellcolor{green!25}55.1\\ 
{Binding} & \cellcolor{green!50}60.7 & \cellcolor{green!50}61.6 & \cellcolor{green!30}58.9 \\
{Control/Raising} & 48.8 & \cellcolor{green!30}59.1 & \cellcolor{green!25}55.6 \\
{Determiner Noun Agreement} & \cellcolor{green!50}65.2 & \cellcolor{green!75}70.9 & \cellcolor{green!60}67.8 \\
{Ellipses} & \cellcolor{green!60}68.6 & \cellcolor{green!60}66.2 & \cellcolor{green!30}57.5\\ 
{Filler Gap} & \cellcolor{green!50}62.4	& \cellcolor{green!60}67.3 & \cellcolor{green!50}62.6 \\ 
{Irregular Forms} & \cellcolor{green!50}61.8 & \cellcolor{green!60}68.2 & \cellcolor{green!75}70.9\\
{Island Effects} & \cellcolor{green!25}51.8	& \cellcolor{green!75}72.7	& \cellcolor{green!25}51.3  \\
{NPI Licensing} & \cellcolor{green!25}53.7 & \cellcolor{green!50}62.6 & \cellcolor{green!25}51.9  \\
{Quantifiers} & \cellcolor{green!30}58.5 & \cellcolor{green!50}62.4 & \cellcolor{green!75}71.7  \\
{Subject Verb Agreement} & \cellcolor{green!25}54.9	& \cellcolor{green!30}57.7 & \cellcolor{green!25}53.8  \\
\bottomrule
\end{tabular}
\caption{Performance of model on BLiMP test suite when trained on different types of input data.}
\label{tab:lang_results}
\end{table*}

We selected five languages at varying distances from English according to their language family and examined how structural dissimilarities with increasing distance from English impact the performance of the model. Figure \ref{fig:real_language_effect} shows the increase in difference between the performance of model trained on English ADS and CDS of the various languages. Our results show negative correlation between the distance of the CDS language from English and the performance of the model, i.e., as the typological distance between L1 and L2 increases, the performance of the model decreases. Based on prior work on transfer errors and typological distance \cite{ringbom}, this decrease in performance could be the result of negative transfer effects, which tend to increase with increase in typological distance between L1 and L2. Among all CDS languages, English CDS performs closest to English ADS (5.13 points behind ADS), suggesting that even within the same language the linguistic differences between ADS and CDS affect model performance (see Table \ref{tab:lang_results}). This is considered as comparisons between other CDS languages and English ADS are made. German shows the next best performance (6.71 points behind English ADS), followed by French (7.27 points behind ADS), Polish (7.57 points behind ADS), Japanese (8.17 points behind ADS), and lastly Indonesian (8.69 points behind ADS). These results confirm our hypothesis that L1s that are structurally closer to L2 (English ADS) perform better, owing to greater degree of positive transfer effects. 

For human language learners, transfer works both ways: sometimes knowledge of parallel structures in the native language facilitate performance in the new language. Other times, there is interference from the native language, resulting in errors. The SLABERT models, similarly, show evidence of both positive and negative transfer. As with human second-language learners, some of the errors we see in SLABERT performance suggest the effect of negative transfer from native [L1] language, while others can be characterized as developmental, in that they are similar to the kinds of errors that even native human speakers will make on their way to learning the target constructions.

\subsection{CDS \& Sources of Errors in Language Learning}
\label{CDS_failure}
Our results show that CDS performs worse than ADS in a majority (9 out of 12) of the grammatical phenomena from the BLiMP test suite (see Figure \ref{fig:results}). We discuss some theoretical explanations for these results.

\textbf{Negation and NPIs:} Child language acquisition research strongly suggests that mastering the full range of negative licensing and anti-licensing contexts takes a long time.  Across languages, detailed acquisition studies find that children do use NPIs with licensing expressions consistently by age 3 or 4 \cite{Tieu2013LogicAG, Lin2015EmergingNT}
 but only with a limited range of negative licensers. Moreover, \citet{Schwab2021OnTA}
showed that, 
even 11 and 12-year-olds, whose language input by that age is entirely ADS, are still in the process of learning some polarity-sensitive expressions. Thus, CDS input alone may not be sufficient for learning the licensing conditions for NPIs. Previous NLP literature also suggests that negation is particularly challenging for language models to learn \cite{Kassner2019NegatedAM, Ettinger2019WhatBI}. Given this, and acquisition studies that have shown that learning licensing conditions for NPIs goes hand-in-hand with learning negation \cite{Wal1996NegativePI}, we expected our model trained on CDS to make \textit{developmental errors} on tests related to NPIs. As discussed in Section \ref{sec:not_expected}, as a Slavic language, Polish also has distinctive constraints on the appearance of NPIs that are the result of competition with grammatical constraints not present in English. In this case, NPI performance is likely subject to both \textit{developmental} errors and \textit{negative transfer} . 

\textbf{Longer Distance Dependencies:} Short and simple sentences are characteristic of CDS. However, it is likely that such utterances do not make ideal training input for LMs to learn long-distance dependencies (LDDs). Consequently, we expect all models trained on CDS data to be negatively impacted on tests that demand long-distance dependency understanding. Island effects, the phenomenon that showed the widest difference in performance compared to ADS-trained (-21.3 points), is one such phenomenon in the BLiMP test suite, requiring long-distance dependency understanding to perform well \cite{island_effects}. Ellipsis and filler-gap structures also depend on LDDs and also suffer from significant decreases in scores compared to ADS (-10.8 and -6.5 points, respectively). This also applies to binding and control/raising phenomena (-2.8 and -3.6 respectively); however, island effects, ellipsis, and filler-gap tests are particularly affected by the model's lack of LDD understanding.

\textbf{Phenomena That Confuse Humans:} \citet{warstadt-etal-2020-blimp-benchmark} report human performance scores which we use to gain an understanding of how our model performs on tests compared to humans. From the reported human performance scores, we observe that not all of the grammatical phenomena in the BLiMP test suite are equally transparent to humans. Human performance on 8 out of 12 phenomena is below 90 points and 3 of those are below 85 points. The lowest is a mean score of 81 for tests on argument structure, where the CDS-trained and ADS-trained models are also seen struggling (rather more seriously) with a mean score of 55.1 and 56.1, respectively. For control/raising, similarly, human performance has a mean score of 84 points while CDS-trained and ADS-trained models have mean scores of 55.6 and 59.1 respectively. We expect CDS to perform poorly on these tests, which are challenging even for people.

\subsection{Negative Transfer}
\label{sec:not_expected}
There are tests where performance of CDS-trained models would be expected to be better given the nature of the phenomena and the characteristics of CDS utterances. However, CDS underperforms compared to ADS even on tests we might expect to be in its wheelhouse. In particular, determiner-noun agreement and subject-verb agreement are the kinds of phenomena that should be easy for the model to learn even from shorter utterances and with relatively small vocabulary size, since they are matters of simple, regular morphology. The results, therefore, are interesting. We hypothesize one reason we do not see good transfer boosts from other-language CDS on these is that patterns of morphology are very language specific.

Looking broadly at the performance of non-English CDS models, we suggest that these results reflect negative cross-linguistic transfer. For example, the distribution of negative polarity items in Polish and many other Slavic languages displays what has been termed the "Bagel problem" \cite{bagel}: because of conflicts with the demands of strict negative concord (in which negation requires multiple elements of an expression must all appear in their negative forms), in Slavic languages, there are NPIs that never appear in what would otherwise be the canonical context of negative polarity licensing, i.e. direct negation \cite{hoeksema}. In this way, language-specific paradigmatic patterns supersede the general correlational relationship between NPIs and their licensing contexts,  producing an opportunity for \textit{negative transfer} and L1 interference effects.

\section{Conclusion}
\label{sec:conclusion}
In this paper, we explore how second language acquisition research and models of second language acquisition can contribute to questions in NLP about the learnability of grammar. Drawing from the previous research on the unique role of child-directed speech (CDS) in language acquisition, we investigate the potential of spontaneously generated CDS to form a special source from which LMs can acquire the structure necessary for first language acquisition. To test sequential second language acquisition in LMs, we introduce SLABERT. The results from our experiments suggest that while positive transfer is a lot more common than negative transfer, negative transfer occurs in LMs just like it occurs in English Second Language (ESL) learners. 
We believe these novel findings call for further research on this front, and suggest that models like SLABERT can provide useful data for testing questions about both language acquisition and typological relationships through patterns of cross-linguistic transfer. To support this, we release our code, novel MAO-CHILDES corpus, and models.

\section{Limitations}
\label{sec:limitations}
Given that many special properties of Child-Directed Speech are not present in text, we would have liked to work on a multimodal dataset, where both visual and speech information would be present. More specifically, we would have liked to test the effect of the following:
\begin{itemize}
    \item Grounding the language models in vision to test the effect of joint attention \cite{Rowe2012ALI, Akhtar2007JointAA}. Joint attention refers to the phenomena where the caregiver's and the child's coordinated attention attention to each other to a third object or an event. 
    \item Child-Directed Speech is known to have special prosodic properties such as higher variability in pitch \cite{Fernald1989ACS, McRoberts1997AccommodationIM, Papouek1991TheMO}, lengthening of vowels and pauses \cite{Albin1996StressedAW, Ratner1986DurationalCW, Fernald1989ACS}, context-specific intonational contours \cite{Katz1996ACO, Papouek1991TheMO, Stern1982IntonationCA}. These properties have been suggested by many researchers to serve as a mechanism for getting the infants attention \cite{cruttenden_1994, Ferguson1977-FERBTA-3, 10.2307/1130938}. This attentive role may be considered to be beneficial for language development in children \cite{Garnica1977-GARSPA-2}. As our models only take text as the input, we were unable to test the relationship the between these properties and language acquisition in neural network based models have. 
    \item Caregivers give a lot of feedback when young children are first producing and acquiring language \cite{Soderstrom2007BeyondBR}. Our current mainstream language models are not interactive. Therefore, it is difficult to incorporate the feedback loop and the test the effect of the same in models' language acquisition. 
    \end{itemize}

As it is, our findings suggest that many of the most important facilitative features of Child-Directed Speech are relevant to precisely those formal and conceptual aspects of language acquisition that are not captured by text-based language models.

In this paper, we have tested the effect of native CDS in L2 acquisition with 5 typologically diverse languages. However, there is enormous scope to test the effect of the same with many more different languages, which may lead to more pointed implications and conclusions than the findings offered here. 

\section{Ethics Statement}
We use publicly available CHILDES data to build our corpora (MAO-CHILDES). Please read more about their terms before using the data.\footnote{\url{https://talkbank.org}} We use the dataset extracted from the CHILDES database only for research purposes and not for commercial reasons. We will release the dataset upon publication under the same license as CHILDES and this is compatible with the license of CHILDES database \cite{CHILDES}. The results of this study are reported on a single run as part of measures taken to avoid computation wastage. We do not foresee any harmful uses of this work. 

\section*{Acknowledgements}
We would like to acknowledge Philip Huebner for clearing our queries regarding the BabyBERTa code-base. We would also like to thank Saujas Vaduguru for helping us improve our initial drafts. We also thank the anonymous reviewers for their feedback on our work.
This work made use of the High Performance Computing Resource in the Core Facility for Advanced Research Computing at Case Western Reserve University.

\bibliography{anthology,custom}

\begin{thebibliography}{67}
\expandafter\ifx\csname natexlab\endcsname\relax\def\natexlab#1{#1}\fi

\bibitem[{Akhtar and Gernsbacher(2007)}]{Akhtar2007JointAA}
Nameera Akhtar and Morton~Ann Gernsbacher. 2007.
\newblock Joint attention and vocabulary development: A critical look.
\newblock \emph{Language and Linguistics Compass}, 1 3:195--207.

\bibitem[{Albin and Echols(1996)}]{Albin1996StressedAW}
Drema~Dial Albin and Catharine~H. Echols. 1996.
\newblock Stressed and word-final syllables in infant-directed speech.
\newblock \emph{Infant Behavior \& Development}, 19:401--418.

\bibitem[{Artetxe et~al.(2018)Artetxe, Labaka, and
  Agirre}]{artetxe-etal-2018-robust}
Mikel Artetxe, Gorka Labaka, and Eneko Agirre. 2018.
\newblock \href {https://doi.org/10.18653/v1/P18-1073} {A robust self-learning
  method for fully unsupervised cross-lingual mappings of word embeddings}.
\newblock In \emph{Proceedings of the 56th Annual Meeting of the Association
  for Computational Linguistics (Volume 1: Long Papers)}, pages 789--798,
  Melbourne, Australia. Association for Computational Linguistics.

\bibitem[{Berzak et~al.(2014)Berzak, Reichart, and
  Katz}]{Berzak2014ReconstructingNL}
Yevgeni Berzak, Roi Reichart, and Boris Katz. 2014.
\newblock Reconstructing native language typology from foreign language usage.
\newblock In \emph{CoNLL}.

\bibitem[{Bock(1986)}]{Bock1986SyntacticPI}
J.~Kathryn Bock. 1986.
\newblock Syntactic persistence in language production.
\newblock \emph{Cognitive Psychology}, 18:355--387.

\bibitem[{Bois(2014)}]{Bois2014TowardsAD}
John W.~Du Bois. 2014.
\newblock Towards a dialogic syntax.
\newblock \emph{Cognitive Linguistics}, 25:359--410.

\bibitem[{Broen(1972)}]{verbalenv}
Patricia Broen. 1972.
\newblock The verbal environment of the language-learning child.
\newblock \emph{Monographs of the American Speech and Hearing Association}, 17.

\bibitem[{Clark(2016)}]{clark_2016}
Eve~V. Clark. 2016.
\newblock \href {https://doi.org/10.1017/CBO9781316534175} {\emph{First
  Language Acquisition}}, 3 edition.
\newblock Cambridge University Press.

\bibitem[{Conneau et~al.(2017)Conneau, Lample, Ranzato, Denoyer, and
  J'egou}]{Conneau2017WordTW}
Alexis Conneau, Guillaume Lample, Marc'Aurelio Ranzato, Ludovic Denoyer, and
  Herv'e J'egou. 2017.
\newblock Word translation without parallel data.
\newblock \emph{ArXiv}, abs/1710.04087.

\bibitem[{Conneau et~al.(2018)Conneau, Rinott, Lample, Williams, Bowman,
  Schwenk, and Stoyanov}]{conneau-etal-2018-xnli}
Alexis Conneau, Ruty Rinott, Guillaume Lample, Adina Williams, Samuel Bowman,
  Holger Schwenk, and Veselin Stoyanov. 2018.
\newblock \href {https://doi.org/10.18653/v1/D18-1269} {{XNLI}: Evaluating
  cross-lingual sentence representations}.
\newblock In \emph{Proceedings of the 2018 Conference on Empirical Methods in
  Natural Language Processing}, pages 2475--2485, Brussels, Belgium.
  Association for Computational Linguistics.

\bibitem[{Cristia(2011)}]{Cristia2011FinegrainedVI}
Alejandrina Cristia. 2011.
\newblock Fine-grained variation in caregivers' /s/ predicts their infants' /s/
  category.
\newblock \emph{The Journal of the Acoustical Society of America}, 129
  5:3271--80.

\bibitem[{Cruttenden(1994)}]{cruttenden_1994}
Alan Cruttenden. 1994.
\newblock \href {https://doi.org/10.1017/CBO9780511620690.008} {\emph{Phonetic
  and prosodic aspects of Baby Talk}}, page 135–152. Cambridge University
  Press.

\bibitem[{Dulay and Burt(1974)}]{Dulay1974ErrorsAS}
Heidi~C. Dulay and Marina~K. Burt. 1974.
\newblock Errors and strategies in child second language acquisition.
\newblock \emph{TESOL Quarterly}, 8:129.

\bibitem[{Ettinger(2019)}]{Ettinger2019WhatBI}
Allyson Ettinger. 2019.
\newblock What bert is not: Lessons from a new suite of psycholinguistic
  diagnostics for language models.
\newblock \emph{Transactions of the Association for Computational Linguistics},
  8:34--48.

\bibitem[{Ferguson(1977)}]{Ferguson1977-FERBTA-3}
Ch~A. Ferguson. 1977.
\newblock Baby talk as a simplified register snow.
\newblock In Catherine~E. Snow and Charles~A. Ferguson, editors, \emph{Talking
  to Children}, pages 209--235. Cambridge University Press.

\bibitem[{Fernald(1989)}]{10.2307/1130938}
Anne Fernald. 1989.
\newblock \href {http://www.jstor.org/stable/1130938} {Intonation and
  communicative intent in mothers' speech to infants: Is the melody the
  message?}
\newblock \emph{Child Development}, 60(6):1497--1510.

\bibitem[{Fernald et~al.(1989)Fernald, Taeschner, Dunn, Papousek,
  de~Boysson-Bardies, and Fukui}]{Fernald1989ACS}
Anne Fernald, Traute Taeschner, Judy Dunn, Mechthild Papousek,
  B{\'e}n{\'e}dicte de~Boysson-Bardies, and I~Fukui. 1989.
\newblock A cross-language study of prosodic modifications in mothers' and
  fathers' speech to preverbal infants.
\newblock \emph{Journal of Child Language}, 16:477 -- 501.

\bibitem[{Gage(1994)}]{10.5555/177910.177914}
Philip Gage. 1994.
\newblock A new algorithm for data compression.
\newblock \emph{C Users J.}, 12(2):23–38.

\bibitem[{Garnica(1977)}]{Garnica1977-GARSPA-2}
Olga~K. Garnica. 1977.
\newblock Some prosodic and paralinguistic features of speech to young
  children.
\newblock In Catherine~E. Snow and Charles~A. Ferguson, editors, \emph{Talking
  to Children}, pages 63--88. Cambridge University Press.

\bibitem[{Gelderloos et~al.(2020)Gelderloos, Chrupała, and
  Alishahi}]{Gelderloos2020LearningTU}
Lieke Gelderloos, Grzegorz Chrupała, and A.~Alishahi. 2020.
\newblock Learning to understand child-directed and adult-directed speech.
\newblock In \emph{Annual Meeting of the Association for Computational
  Linguistics}.

\bibitem[{Gries(2005)}]{Gries2005SyntacticPA}
Stefan~Th. Gries. 2005.
\newblock Syntactic priming: A corpus-based approach.
\newblock \emph{Journal of Psycholinguistic Research}, 34:365--399.

\bibitem[{Haggan(2002)}]{HAGGAN200217}
Madeline Haggan. 2002.
\newblock \href {https://doi.org/https://doi.org/10.1016/S0388-0001(00)00044-9}
  {Self-reports and self-delusion regarding the use of motherese: implications
  from kuwaiti adults}.
\newblock \emph{Language Sciences}, 24(1):17--28.

\bibitem[{Hayes and Ahrens(1988)}]{hayes_ahrens_1988}
Donald~P. Hayes and Margaret~G. Ahrens. 1988.
\newblock \href {https://doi.org/10.1017/S0305000900012411} {Vocabulary
  simplification for children: a special case of ‘motherese’?}
\newblock \emph{Journal of Child Language}, 15(2):395–410.

\bibitem[{Hochreiter and Schmidhuber(1997)}]{lstm}
Sepp Hochreiter and Jürgen Schmidhuber. 1997.
\newblock \href {https://doi.org/10.1162/neco.1997.9.8.1735} {Long short-term
  memory}.
\newblock \emph{Neural computation}, 9:1735--80.

\bibitem[{Hoeksema(2012)}]{hoeksema}
Jack Hoeksema. 2012.
\newblock On the natural history of negative polarity items.
\newblock \emph{Linguistic Analysis}, 44:3--33.

\bibitem[{Hornstein et~al.(2005)Hornstein, Nunes, and
  Grohmann}]{hornstein_nunes_grohmann_2005}
Norbert Hornstein, Jairo Nunes, and Kleanthes~K. Grohmann. 2005.
\newblock \href {https://doi.org/10.1017/CBO9780511840678} {\emph{Understanding
  Minimalism}}.
\newblock Cambridge Textbooks in Linguistics. Cambridge University Press.

\bibitem[{Huebner et~al.(2021)Huebner, Sulem, Cynthia, and
  Roth}]{huebner-etal-2021-babyberta}
Philip~A. Huebner, Elior Sulem, Fisher Cynthia, and Dan Roth. 2021.
\newblock \href {https://doi.org/10.18653/v1/2021.conll-1.49} {{B}aby{BERT}a:
  Learning more grammar with small-scale child-directed language}.
\newblock In \emph{Proceedings of the 25th Conference on Computational Natural
  Language Learning}, pages 624--646, Online. Association for Computational
  Linguistics.

\bibitem[{Huebner and Willits(2021)}]{inbook}
Philip~A. Huebner and Jon~A. Willits. 2021.
\newblock \href {https://doi.org/https://doi.org/10.1016/bs.plm.2021.08.002}
  {Chapter eight - using lexical context to discover the noun category: Younger
  children have it easier}.
\newblock In Kara~D. Federmeier and Lili Sahakyan, editors, \emph{The Context
  of Cognition: Emerging Perspectives}, volume~75 of \emph{Psychology of
  Learning and Motivation}, pages 279--331. Academic Press.

\bibitem[{Jarvis and Pavlenko(2007)}]{Jarvis2007CrosslinguisticII}
Scott Jarvis and Aneta Pavlenko. 2007.
\newblock Crosslinguistic influence in language and cognition.

\bibitem[{Kassner and Sch{\"u}tze(2019)}]{Kassner2019NegatedAM}
Nora Kassner and Hinrich Sch{\"u}tze. 2019.
\newblock Negated and misprimed probes for pretrained language models: Birds
  can talk, but cannot fly.
\newblock In \emph{Annual Meeting of the Association for Computational
  Linguistics}.

\bibitem[{Katz et~al.(1996)Katz, Cohn, and Moore}]{Katz1996ACO}
Gary~S. Katz, Jeffrey~F. Cohn, and Christopher~A. Moore. 1996.
\newblock A combination of vocal fo dynamic and summary features discriminates
  between three pragmatic categories of infant-directed speech.
\newblock \emph{Child development}, 67 1:205--17.

\bibitem[{Kirchhoff and Schimmel(2003)}]{Kirchhoff2003StatisticalPO}
Katrin Kirchhoff and Steven~M. Schimmel. 2003.
\newblock Statistical properties of infant-directed versus adult-directed
  speech: insights from speech recognition.
\newblock \emph{The Journal of the Acoustical Society of America}, 117 4 Pt
  1:2238--46.

\bibitem[{K{\"o}ymen and Kyratzis(2014)}]{Kymen2014DialogicSA}
Bahar K{\"o}ymen and Amy Kyratzis. 2014.
\newblock Dialogic syntax and complement constructions in toddlers' peer
  interactions.
\newblock \emph{Cognitive Linguistics}, 25:497 -- 521.

\bibitem[{Küntay and Slobin(2002)}]{turkishlangacq}
Aylin Küntay and Dan Slobin. 2002.
\newblock Putting interaction back into child language: Examples from turkish.
\newblock \emph{Psychology of Language and Communication, v.6 (2002)}, 6.

\bibitem[{Lin et~al.(2015)Lin, Weerman, and Zeijlstra}]{Lin2015EmergingNT}
Jing Lin, F.~P. Weerman, and Hedde Zeijlstra. 2015.
\newblock Emerging npis: The acquisition of dutch hoeven ‘need’.
\newblock \emph{The Linguistic Review}, 32:333 -- 374.

\bibitem[{Liu et~al.(2019)Liu, Ott, Goyal, Du, Joshi, Chen, Levy, Lewis,
  Zettlemoyer, and Stoyanov}]{Liu2019RoBERTaAR}
Yinhan Liu, Myle Ott, Naman Goyal, Jingfei Du, Mandar Joshi, Danqi Chen, Omer
  Levy, Mike Lewis, Luke Zettlemoyer, and Veselin Stoyanov. 2019.
\newblock Roberta: A robustly optimized bert pretraining approach.
\newblock \emph{ArXiv}, abs/1907.11692.

\bibitem[{Ma et~al.(2011)Ma, Golinkoff, Houston, and
  Hirsh-Pasek}]{Ma2011WordLI}
Weiyi Ma, Roberta~Michnick Golinkoff, Derek~M. Houston, and Kathy Hirsh-Pasek.
  2011.
\newblock Word learning in infant- and adult-directed speech.
\newblock \emph{Language Learning and Development}, 7:185 -- 201.

\bibitem[{Macwhinney(2000)}]{CHILDES}
Brian Macwhinney. 2000.
\newblock \href {https://doi.org/10.1162/coli.2000.26.4.657} {The childes
  project: Tools for analyzing talk (third edition): Volume i: Transcription
  format and programs, volume ii: The database}.
\newblock \emph{Computational Linguistics - COLI}, 26:657--657.

\bibitem[{McRoberts and Best(1997)}]{McRoberts1997AccommodationIM}
Gerald McRoberts and Catherine~T. Best. 1997.
\newblock Accommodation in mean f0 during mother–infant and father–infant
  vocal interactions: a longitudinal case study.
\newblock \emph{Journal of Child Language}, 24:719 -- 736.

\bibitem[{Mintz(2003)}]{Mintz2003FrequentFA}
Toben~H. Mintz. 2003.
\newblock Frequent frames as a cue for grammatical categories in child directed
  speech.
\newblock \emph{Cognition}, 90:91--117.

\bibitem[{Nelson et~al.(1986)Nelson, Hirsh-Pasek, Jusczyk, and
  Cassidy}]{KemlerNelson1986HowTP}
Deborah G~Kemler Nelson, Kathy Hirsh-Pasek, Peter~W. Jusczyk, and
  Kimberly~Wright Cassidy. 1986.
\newblock How the prosodic cues in motherese might assist language learning.
\newblock \emph{Journal of Child Language}, 16:55 -- 68.

\bibitem[{Papadimitriou and Jurafsky(2020)}]{music}
Isabel Papadimitriou and Dan Jurafsky. 2020.
\newblock \href {https://doi.org/10.18653/v1/2020.emnlp-main.554} {{L}earning
  {M}usic {H}elps {Y}ou {R}ead: {U}sing transfer to study linguistic structure
  in language models}.
\newblock In \emph{Proceedings of the 2020 Conference on Empirical Methods in
  Natural Language Processing (EMNLP)}, pages 6829--6839, Online. Association
  for Computational Linguistics.

\bibitem[{Papousek et~al.(1991)Papousek, Papousek, and
  Symmes}]{Papouek1991TheMO}
Mechthild Papousek, Hanǔs Papousek, and David~T Symmes. 1991.
\newblock The meanings of melodies in motherese in tone and stress languages.
\newblock \emph{Infant Behavior \& Development}, 14:415--440.

\bibitem[{Pereltsvaig(2006)}]{bagel}
Asya Pereltsvaig. 2006.
\newblock \href {https://doi.org/10.1007/s11049-005-3820-z} {Small nominals}.
\newblock \emph{Natural Language and Linguistic Theory}, 24:433--500.

\bibitem[{Pickering and Ferreira(2008)}]{Pickering2008StructuralPA}
Martin~John Pickering and Victor~S. Ferreira. 2008.
\newblock Structural priming: a critical review.
\newblock \emph{Psychological bulletin}, 134 3:427--59.

\bibitem[{Pinker(1995)}]{pinker_1995}
Steven Pinker. 1995.
\newblock \emph{The Language Instinct}.
\newblock PENGUIN.

\bibitem[{Ratner(1986)}]{Ratner1986DurationalCW}
Nan~Bernstein Ratner. 1986.
\newblock Durational cues which mark clause boundaries in mother–child
  speech.
\newblock \emph{Journal of Phonetics}, 14:303--309.

\bibitem[{Ringbom(2006)}]{ringbom}
Håkan Ringbom. 2006.
\newblock \href {https://doi.org/doi:10.21832/9781853599361}
  {\emph{Cross-linguistic Similarity in Foreign Language Learning}}.
\newblock Multilingual Matters, Bristol, Blue Ridge Summit.

\bibitem[{Rowe(2012)}]{Rowe2012ALI}
Meredith~L. Rowe. 2012.
\newblock A longitudinal investigation of the role of quantity and quality of
  child-directed speech in vocabulary development.
\newblock \emph{Child development}, 83 5:1762--74.

\bibitem[{Ruder et~al.(2017)Ruder, Vulic, and S{\o}gaard}]{Ruder2017ASO}
Sebastian Ruder, Ivan Vulic, and Anders S{\o}gaard. 2017.
\newblock A survey of cross-lingual word embedding models.
\newblock \emph{J. Artif. Intell. Res.}, 65:569--631.

\bibitem[{Sanchez et~al.(2019)Sanchez, Meylan, Braginsky, Macdonald, Yurovsky,
  and Frank}]{childes-db}
Alessandro Sanchez, Stephan Meylan, Mika Braginsky, Kyle Macdonald, Daniel
  Yurovsky, and Michael Frank. 2019.
\newblock \href {https://doi.org/10.3758/s13428-018-1176-7} {childes-db: A
  flexible and reproducible interface to the child language data exchange
  system}.
\newblock \emph{Behavior Research Methods}, 51.

\bibitem[{Saxton(2009)}]{Saxton2009TheIO}
Matthew~L. Saxton. 2009.
\newblock The inevitability of child directed speech.

\bibitem[{Schwab et~al.(2021)Schwab, Liu, and Mueller}]{Schwab2021OnTA}
Juliane Schwab, Mingya Liu, and Jutta~L. Mueller. 2021.
\newblock On the acquisition of polarity items: 11- to 12-year-olds'
  comprehension of german npis and ppis.
\newblock \emph{Journal of Psycholinguistic Research}, 50:1487 -- 1509.

\bibitem[{Snow(1972)}]{10.2307/1127555}
Catherine~E. Snow. 1972.
\newblock \href {http://www.jstor.org/stable/1127555} {Mothers' speech to
  children learning language}.
\newblock \emph{Child Development}, 43(2):549--565.

\bibitem[{Soderstrom(2007)}]{Soderstrom2007BeyondBR}
Melanie Soderstrom. 2007.
\newblock Beyond babytalk: Re-evaluating the nature and content of speech input
  to preverbal infants.
\newblock \emph{Developmental Review}, 27:501--532.

\bibitem[{Soderstrom et~al.(2008)Soderstrom, Blossom, Foygel, and
  Morgan}]{Soderstrom2008AcousticalCA}
Melanie Soderstrom, Megan~Stratton Blossom, Rina Foygel, and James~L. Morgan.
  2008.
\newblock Acoustical cues and grammatical units in speech to two preverbal
  infants*.
\newblock \emph{Journal of Child Language}, 35:869 -- 902.

\bibitem[{Sprouse and Hornstein(2013)}]{island_effects}
Jon Sprouse and Norbert Hornstein. 2013.
\newblock \href {https://doi.org/10.1017/CBO9781139035309.001}
  {\emph{Experimental syntax and island effects: Toward a comprehensive theory
  of islands}}, page 1–18. Cambridge University Press.

\bibitem[{Stern et~al.(1982)Stern, Spieker, and
  Mackain}]{Stern1982IntonationCA}
Daniel~N. Stern, Susan~J. Spieker, and K.~Mackain. 1982.
\newblock Intonation contours as signals in maternal speech to prelinguistic
  infants.
\newblock \emph{Developmental Psychology}, 18:727--735.

\bibitem[{Thiessen et~al.(2005)Thiessen, Hill, and Saffran}]{thiessen2005}
Erik~D. Thiessen, Emily~A. Hill, and Jenny~R. Saffran. 2005.
\newblock \href {https://doi.org/https://doi.org/10.1207/s15327078in0701\_5}
  {Infant-directed speech facilitates word segmentation}.
\newblock \emph{Infancy}, 7(1):53--71.

\bibitem[{Tieu(2013)}]{Tieu2013LogicAG}
Lyn Tieu. 2013.
\newblock Logic and grammar in child language: How children acquire the
  semantics of polarity sensitivity.

\bibitem[{van~der Wal(1996)}]{Wal1996NegativePI}
S.~van~der Wal. 1996.
\newblock Negative polarity items and negation: Tandem acquisition.

\bibitem[{Vaswani et~al.(2017)Vaswani, Shazeer, Parmar, Uszkoreit, Jones,
  Gomez, Kaiser, and Polosukhin}]{Vaswani2017AttentionIA}
Ashish Vaswani, Noam~M. Shazeer, Niki Parmar, Jakob Uszkoreit, Llion Jones,
  Aidan~N. Gomez, Lukasz Kaiser, and Illia Polosukhin. 2017.
\newblock Attention is all you need.
\newblock In \emph{NIPS}.

\bibitem[{Veneziano and Parisse(2010)}]{verb}
Edy Veneziano and Christophe Parisse. 2010.
\newblock \href {https://doi.org/10.1177/0142723710379785} {The acquisition of
  early verbs in french: Assessing the role of conversation and of
  child-directed speech}.
\newblock \emph{International Conference on Infant Studies 2010}, 30.

\bibitem[{Warstadt et~al.(2020)Warstadt, Parrish, Liu, Mohananey, Peng, Wang,
  and Bowman}]{warstadt-etal-2020-blimp-benchmark}
Alex Warstadt, Alicia Parrish, Haokun Liu, Anhad Mohananey, Wei Peng, Sheng-Fu
  Wang, and Samuel~R. Bowman. 2020.
\newblock \href {https://doi.org/10.1162/tacl_a_00321} {{BL}i{MP}: The
  benchmark of linguistic minimal pairs for {E}nglish}.
\newblock \emph{Transactions of the Association for Computational Linguistics},
  8:377--392.

\bibitem[{Wolf et~al.(2019)Wolf, Debut, Sanh, Chaumond, Delangue, Moi, Cistac,
  Rault, Louf, Funtowicz, and Brew}]{Wolf2019HuggingFacesTS}
Thomas Wolf, Lysandre Debut, Victor Sanh, Julien Chaumond, Clement Delangue,
  Anthony Moi, Pierric Cistac, Tim Rault, R{\'e}mi Louf, Morgan Funtowicz, and
  Jamie Brew. 2019.
\newblock Huggingface's transformers: State-of-the-art natural language
  processing.
\newblock \emph{ArXiv}, abs/1910.03771.

\bibitem[{Wu et~al.(2019)Wu, Conneau, Li, Zettlemoyer, and
  Stoyanov}]{Wu2019EmergingCS}
Shijie Wu, Alexis Conneau, Haoran Li, Luke Zettlemoyer, and Veselin Stoyanov.
  2019.
\newblock Emerging cross-lingual structure in pretrained language models.
\newblock \emph{ArXiv}, abs/1911.01464.

\bibitem[{Wu and Dredze(2019)}]{wu-dredze-2019-beto}
Shijie Wu and Mark Dredze. 2019.
\newblock \href {https://doi.org/10.18653/v1/D19-1077} {Beto, bentz, becas: The
  surprising cross-lingual effectiveness of {BERT}}.
\newblock In \emph{Proceedings of the 2019 Conference on Empirical Methods in
  Natural Language Processing and the 9th International Joint Conference on
  Natural Language Processing (EMNLP-IJCNLP)}, pages 833--844, Hong Kong,
  China. Association for Computational Linguistics.

\end{thebibliography}
\bibliographystyle{acl_natbib}

\appendix

\section{Appendix}
\label{sec:appendix}

\begin{figure*}[t]
\centering
\includegraphics[width=\textwidth]{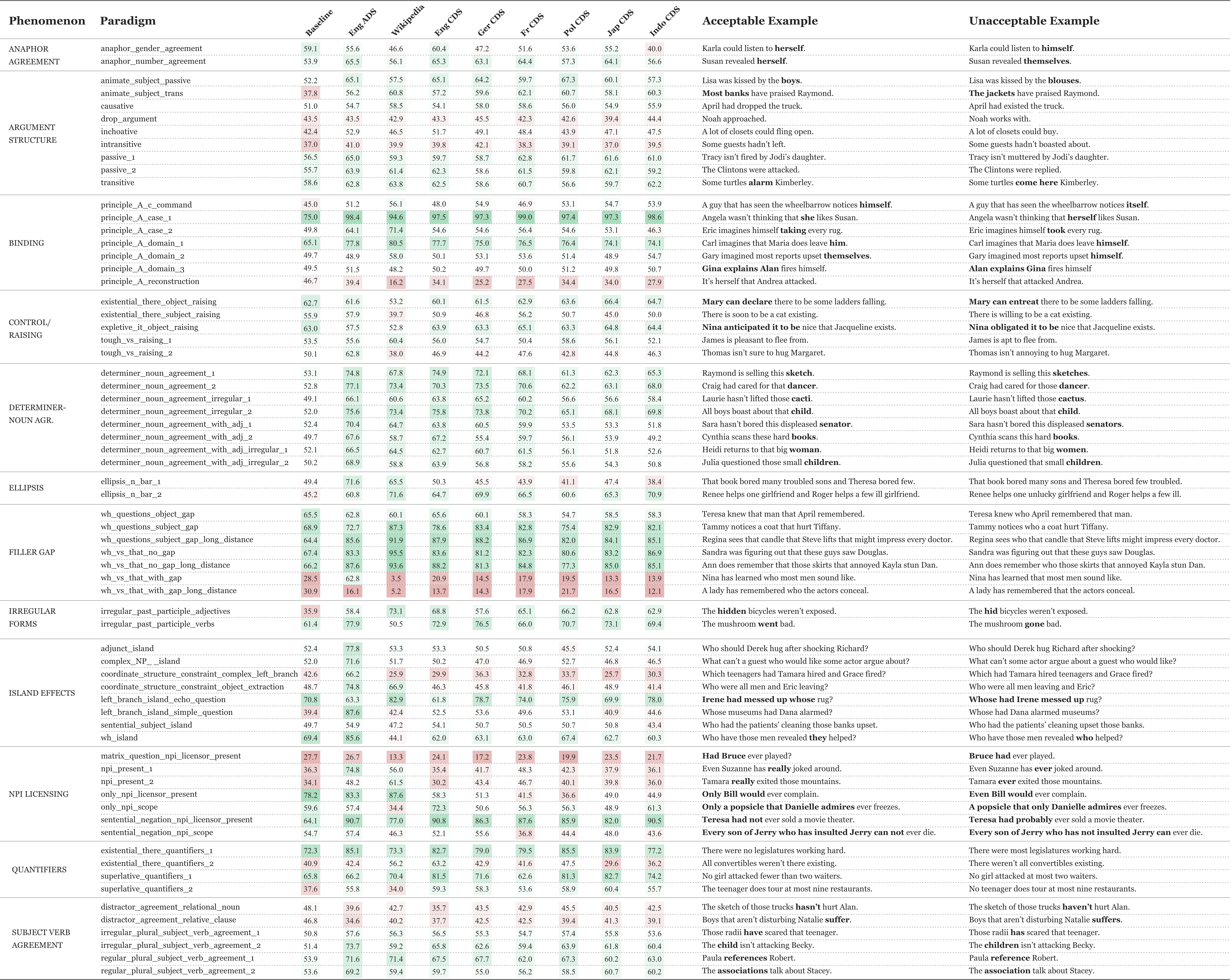}
\caption{Performance of model on all 67 paradigms in BLiMP test suite along with examples of minimal pairs}
\label{fig:results_tab}
\end{figure*}

\subsection{Implementation Details}
\label{sec:imp_details}

We conduct our experiments using BabyBERTa \cite{huebner-etal-2021-babyberta}, a RoBERTa-based model \cite{Liu2019RoBERTaAR}, with 8 hidden layers, 8 attention heads, and dimensionality of the encoder layer being 256, dimensionality of the intermediate or the feed-forward layer in the transfer based encoder being 1024. We train this model with a learning rate of 1e-4, batch size of 16 and limit the maximum sequence length to 128. This model is trained for 10 epochs with max step size of 260. We train this on a single V100 GPU. To tokenize the words we use Byte Pair Encoder (BPE) \cite{10.5555/177910.177914} based tokenizer with vocabulary size set to 52,000 and minimum frequency set to 2. The rest of the hyperparameters are set to their default settings in the Transformers library \cite{Wolf2019HuggingFacesTS}.

\subsection{Comprehensive Results}
\label{sec:comp_results}
 
Figure \ref{fig:results_tab} illustrates the organization of the BLiMP test suite and the performance of all models along with examples of minimal pairs from each of the 67 paradigms.

\subsubsection{Organization of BLiMP}
BLiMP consists of 67 minimal pair paradigms grouped into 12 distinct linguistic phenomena: anaphor agreement, argument structure, binding, control/raising, determiner-noun agreement, ellipsis, filler gap, irregular forms, island effects, NPI licensing, quantifiers, and subject-verb agreement. Each paradigm comprises 1,000 sentence pairs in English and isolates specific phenomenon in syntax, morphology, or semantics. A complete description of each linguistic phenomenon and finer details of the test suite can be found in \citet{warstadt-etal-2020-blimp-benchmark}.

\subsubsection{Models}
A total of 9 models are used in our study. (1) The Random Baseline model that is specifically trained such that it acquires no grammatical structure from the training data and only acquires English vocabulary (2) the Wikipedia-4 model that is trained on scripted ADS English data (3) the English ADS model that is trained on transcriptions of spontaneous, conversational speech in English (4) the English CDS model (5) the German CDS model (6) the French CDS model (7) the Polish CDS model (8) the Japanese CDS model (9) the Indonesian CDS model, where models 4 through 9 are trained on conversational CDS data from 6 different languages.

\end{document}